\documentclass[smallcondensed]{svjour3}

\usepackage{url,amsmath,amssymb,amsfonts}
\usepackage{algorithmic}
\usepackage{graphicx,multirow,comment}
\usepackage{float} 
\usepackage{xcolor}
\graphicspath{{figs/}}
\usepackage{rotating}

\begin{document}

\title{Reflective-Net: Learning from Explanations}
\subtitle{}

\author{Johannes Schneider \and Michalis Vlachos }
\institute{J. Schneider \at Institute of Information Systems\\ University of Liechtenstein, Liechtenstein \\
\email{johannes.schneider@uni.li}
\and M. Vlachos \at HEC Lausanne, Department of Information Systems\\ University of Lausanne, Switzerland  }




\maketitle

\begin{abstract} 

We examine whether data generated by explanation techniques, which promote a process of self-reflection, can improve classifier performance. Our work is based on the idea that humans have the ability to make quick, intuitive decisions as well as to reflect on their own thinking and learn from explanations. To the best of our knowledge, this is the first time that the potential of mimicking this process by using explanations generated by explainability methods has been explored. We found that combining explanations with traditional labeled data leads to significant improvements in classification accuracy and training efficiency across multiple image classification datasets and convolutional neural network architectures. It is worth noting that during training, we not only used explanations for the correct or predicted class, but also for other classes. This serves multiple purposes, including allowing for reflection on potential outcomes and enriching the data through augmentation.
\end{abstract}

\section{Introduction} 
Self-reflection can be defined as ``The capacity of humans to exercise introspection and to attempt to learn more about their fundamental nature and essence.''[Wikipedia]
Reflective thinking is an essential process that has led to numerous notable achievements in literature and science. However, it has not been a concept previously applied in machine learning. The emerging field of explainable artificial intelligence (XAI) \cite{mes21} has introduced  techniques generating explanations. \emph{Explanations} support the understanding of machine learning models. They can provide a wealth of information on model behavior, as multiple explanations can be generated for each input. Humans have primarily used this data to understand decisions made by AI systems. In the field of interactive machine learning, feedback loops have been established, where explanations originating from XAI techniques are presented to humans, who can adjust them and provide feedback to the machine learning system~\cite{schm20}. However, there has been no form of self-introspection in the learning process to date.

In this work, we aim to leverage the vast data generated through explanations to improve machine learning models. Our approach can be seen as a form of data enhancement, where the raw inputs are enhanced through the inclusion of explanations. The reflective process also allows for new opportunities for data augmentation, where the ground truth and the explanation of the correct class serve as the original input. Explanations of random classes can be used for augmentation. In some sense, this work shows how machine learning can ``mine'' its own data generated in the form of explanations. Thus, here, \emph{learning from explanations} refers to the process of including explanations originating from (adjusted) XAI techniques in the learning process in addition to training data to improve a machine learning system.

\begin{figure}[!h]
  \centering{  \includegraphics[width=0.6\linewidth]{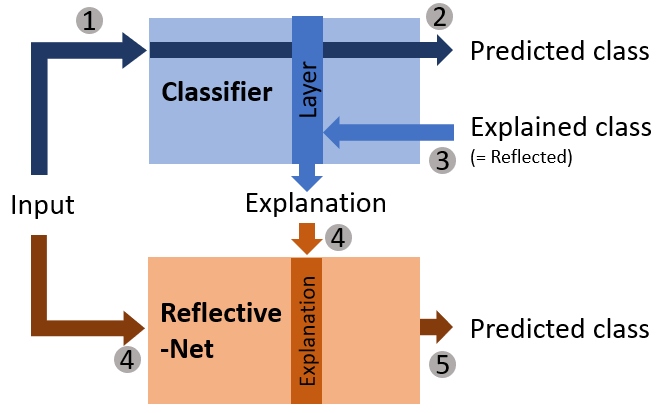}}
  \caption{Reflective-Net: After a first classification (Step 1-2), a possible decision for the input is explained using a backward pass up to a specific layer (Step 3). Then, the explanation and input (Step 4) are used to get a second classification.} \label{fig:basicRef}
 \end{figure}

Our approach also builds on basic concepts of human reasoning, though we do not claim to fully replicate actual human reasoning.  In his book ``Thinking, Fast and Slow'',  Nobel laureate Daniel Kahneman proposes the idea that humans have two different modes of thinking: (i) a fast, unconscious, instinctive system for tasks such as determining the relative distance of object and (ii) a slow, conscious, deliberative system, for tasks such as focusing on a specific person in a crowded, noisy environment. Inspired by this idea, we extend classical inference in deep learning, which is based on a single (fast) forward pass, to include a more reflective, slower inference process using explanations. We propagate from a class to explain, i.e., the final layer, back to a specific layer of the neural network using an adjusted version of the well-known explainability method ``GradCAM''\cite{selvaraju2017grad}. We then use the explanation and the input together to make a final decision, as shown in Figure \ref{fig:basicRef}. We refer to this process of explaining a \textit{fast} prediction and using the original input and explanation to make a final decision as ``reflecting.''

\begin{figure*}[!ht]
  \centering{  \includegraphics[width=\linewidth]{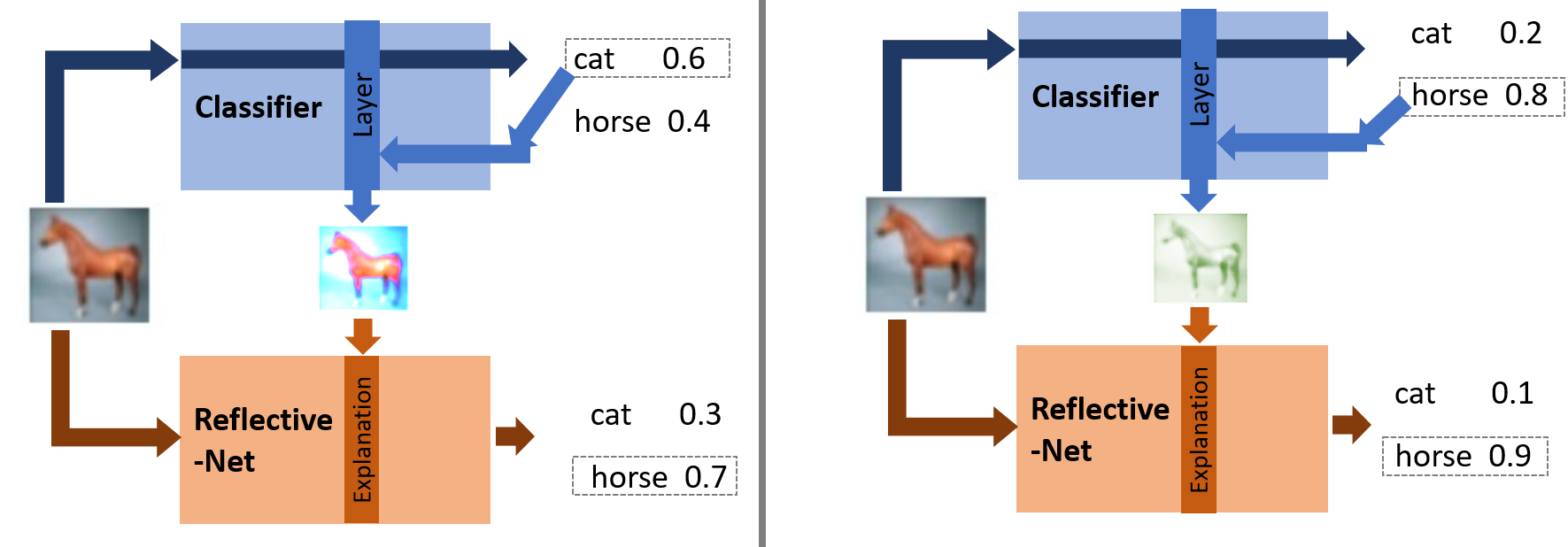}}
  \caption{Reflective-Net: Learning from explanations of correct and incorrect predictions. Predicted classes have grey boundaries. Reflective thinking based on either incorrect (left panel) or correct predictions (right panel) leads to better outcomes.} \label{fig:refNet}
 \end{figure*}

The reflective process includes a form of deliberation, in which we evaluate multiple options by testing or making assumptions about the class of an input that may differ from the ground truth or predicted class. So, reflecting can include explanations for different outcomes than the actual prediction or the ground truth. Figure \ref{fig:refNet} illustrates the use of an explanation for input based on the correct class and an incorrect class.

In humans, self-reflection can be distinguished from non-self-reflection.  While self-reflection is typically a lengthy and iterative process, we use the term ``self-reflection'' in our study to refer to a single backward pass. Self-reflection is an evolutionary process that develops over time and depends on skills that are acquired after mastering basic perceptual skills, such as object recognition. In our study, the ``reflective'' network is also an evolution of the classifier used to generate explanations in Figure \ref{fig:basicRef}. We investigate both ``self'' vs. ``non-self'' reflection in our study, meaning that the reflective-net can be the result of ``fine-tuning'' using explanations of an existing classifier, or it can be trained from scratch using explanations from an existing classifier with randomly initialized parameters.

Our primary goal in using reflection is to improve the learning process. During training, we consider different potential outcomes, even if they are unlikely, through reflection. In contrast, during inference, we only rely on explanations of actual predictions. Therefore, we view reflection as a tool that primarily supports the generalization of the network. In our empirical analysis, we also evaluate the performance of the network using an oracle that provides explanations for the correct class.

Two of our findings are that (i) training using explanations based on different prediction outcomes leads to better generalization and (ii) using the ground truth explanation to make a final decision results in very large accuracy improvements, while training on the predicted outcome still yields notable improvements. Our \textbf{contributions} include:
 
\begin{itemize}
    \item Demonstrating how to learn from explanations by modifying the well-known XAI method GradCAM to obtain more informative explanations, and using the idea of enhancing inputs with explanations for different prediction outcomes (classes).
    \item Empirically showing the benefits of our method and the impact of various design options. Our reflective network using explanations outperforms non-reflective classifiers in terms of accuracy.
\end{itemize}

\section{Reflective Networks}
We implement the ``reflection process'' of a neural network as follows. First, a classifier makes an initial prediction for an input. Then, it reasons upon this prediction or another possible outcome, yielding an explanation (Figure \ref{fig:basicRef}). The explanation and the input sample can then be used by the same or a different network to produce a second prediction. For humans, reflection also involves considering and envisioning different scenarios or outcomes, or predictions. For example, the network may be given the explanation of the correct class even though it predicted another class (right panel in Figure \ref{fig:refNet}). Our results show diversity of explanations per sample is a must. Training a system with just one explanation for an input, i.e.,  always the correct or the actual prediction yields limited or no improvement. Therefore, it is essential to consider different outcomes during training to prevent the network from relying too heavily on explanations (while still using the original input) and  to help the network identify correct and incorrect explanations for given inputs and how to transform them into the correct outcome. Additionally, we can control the amount of information per explantion. Traditionally, gradient-based attriubtion methods like GradCAM methods highlight pixels in support (or in contradiction) to a prediction. Thus, they consist of one channel, i.e., one number pixel. We use more informative explanations. Our explanations have both a spatial extent and a ``depth'' of multiple channels. In addition, we aim to provide explanations at a higher level of abstraction, or more semantically meaningful features of an object, rather than individual pixels. As a result, our explanations are computed for intermediate layers rather than the input and have a different shape than the input. In contrast to GradCAM, we do not reshape the explanations to the input shape through up-sampling and aggregation of channels.

To summarize, our explanations focus on more semantically meaningful, or intermediate, layers using a "semantic" or depth dimension. We begin by describing the simplest system architecture for leveraging explanations, then we elaborate on the non-reflective classifier, how to compute explanations, and how to incorporate them into a classifier. In Section \ref{sec:opt}, we discuss various design decisions.

\subsection{Base Architecture}

The overall goal is to compute explanations for an input to a trained classifier. There are design decisions to be made about the level of abstraction for the explanations, such as whether they should be based on pixels or higher-level concepts. For example, should the answer to the question ``Why is this a car?" be based on low-level details like the presence of certain pixels or on more semantically meaningful features like the presence of two visible tires and a red fender and front door? Another design decision is about which decisions (classes) the explanations should be based on, such as only the predicted class or all possible classes.

Once the explanations have been computed, they and the explained sample are used as input for the same (or a different) classifier. An important question is how to incorporate the explanations into the classifier. In the following sections, we will provide details about the base architecture and discuss these design decisions.

\begin{figure*}
 \centering{  \includegraphics[width=0.9\linewidth]{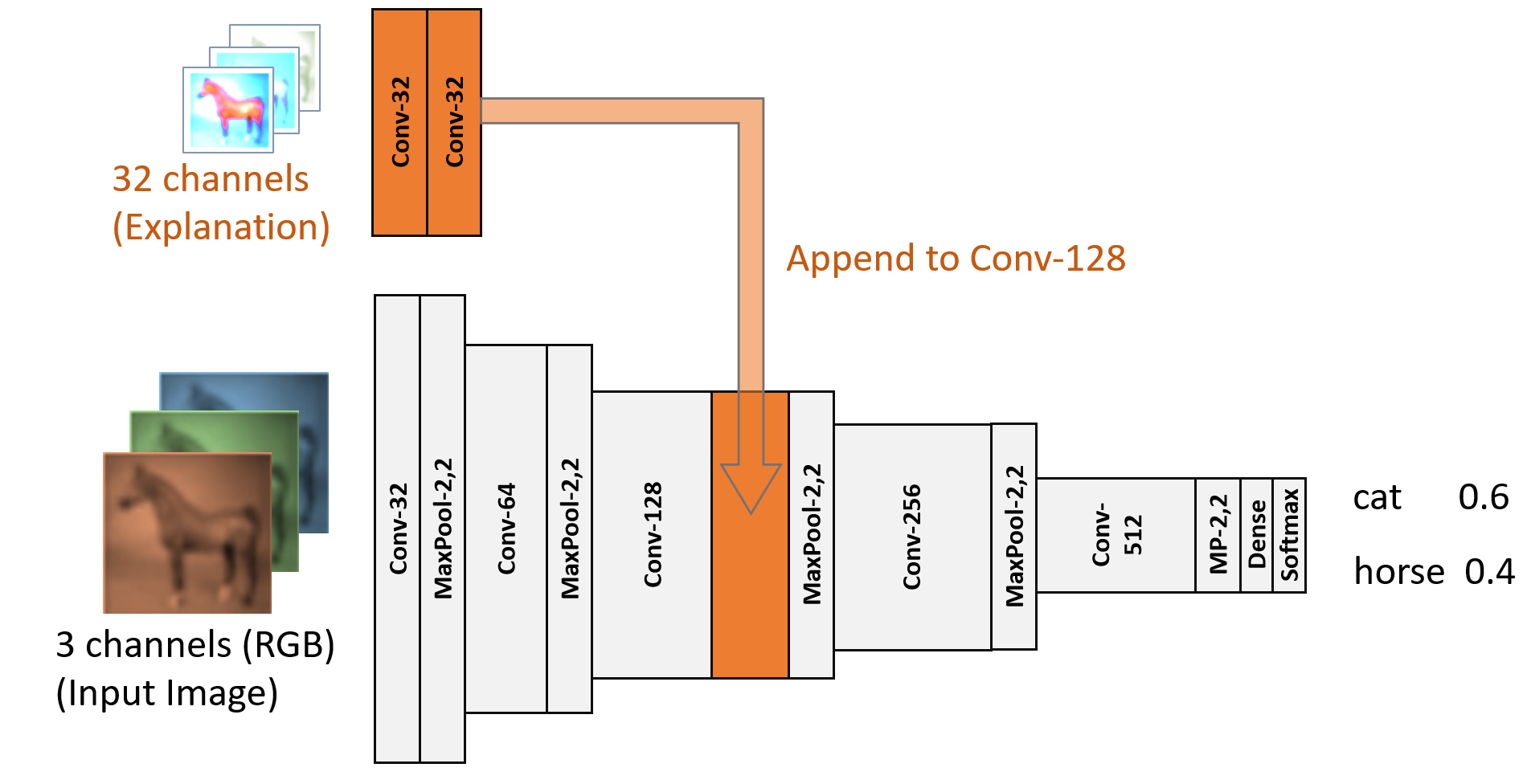}}
 \caption{Reflective-Net illustrated for VGG; A VGG network (grey boxes) is extended to a network using explanations. Explanations pass through two conv-layers (orange) and the output is appended; \scriptsize{(each conv layer is followed by a batchnorm and a relu layer)}} \label{fig:arch}
\end{figure*}

\medskip
\noindent\emph{Non-reflective classifier:} We assume the existence of a trained classifier $C_O$ using an arbitrary architecture. We evaluate the performance of two specific types of architectures, ResNet and VGG. For illustration, we will use a VGG-style convolutional neural network as an example, which can be seen in Table \ref{tab:arch} and Figure \ref{fig:arch} without the orange-colored layers. For this classifier $C_O$, explanations are computed. 

\begin{table}[h] 	
 	   	\begin{center}
 		\setlength\tabcolsep{2.5pt}
	\centering
 		\begin{tabular}{|l|l| l | l| l|l| }\hline
 			\multicolumn{3}{|c|}{\footnotesize{ResNet variant\cite{he16d}}} & \multicolumn{2}{c|}{\footnotesize{VGG variant\cite{sim14}}} & Explained \\ \cline{1-5}
		     Block &Type/Stride & Filter Shape &Type/Stride& Filter Shape & Layer $L$ \\  \hline
 			  - & C/s1     & $3\tiny{\times} 3 \tiny{\times} 3 \tiny{\times} 64$ & C/MP     & $3\tiny{\times} 3 \tiny{\times} 3 \tiny{\times} 32$ &  \\ \hline
 			  \multirow{2}{*}{1} & C/s1     & $3\tiny{\times} 3 \tiny{\times} 64\tiny{\times} 64$  & C/MP & $3\tiny{\times} 3 \tiny{\times} 32 \tiny{\times} 64$& Low   \\ \cline{2-6}
 		      & C/s1     & $3\tiny{\times} 3 \tiny{\times} 64 \tiny{\times} 128$ &   C     &$3\tiny{\times} 3 \tiny{\times} 64 \tiny{\times} 128$ & \\ \hline 
 			\multirow{2}{*}{2} & C/s2     & $3\tiny{\times} 3 \tiny{\times} 128 \tiny{\times} 128$ && &  \\ \cline{2-6}
 			  & C/s1     & $3\tiny{\times} 3 \tiny{\times} 128 \tiny{\times} 256$ &  C/MP     &$3\tiny{\times} 3 \tiny{\times} 128 \tiny{\times} 128$  & Middle \\ \hline
 			 \multirow{2}{*}{3} &C/s2     & $3\tiny{\times} 3 \tiny{\times} 256\tiny{\times} 256$ & C &  $3\tiny{\times} 3 \tiny{\times} 128 \tiny{\times} 256$  &  \\ \cline{2-6}
 			  &C/s1     & $3\tiny{\times} 3 \tiny{\times} 256 \tiny{\times} 512$ & C & $3\tiny{\times} 3 \tiny{\times} 256 \tiny{\times} 256$ & High\\ \hline
 			  \multirow{2}{*}{4} &C/s2     & $3\tiny{\times} 3 \tiny{\times} 512\tiny{\times} 512$ &   C/MP     & $3\tiny{\times} 3 \tiny{\times} 256 \tiny{\times} 512$ &\\ \cline{2-6}
 			   &C/s1     & $3\tiny{\times} 3 \tiny{\times} 512 \tiny{\times} 512$ &  C     & $3\tiny{\times} 3 \tiny{\times} 512 \tiny{\times} 512$  & \\ \hline  			   
 			  -&FC/s1 & $512 \tiny{\times} $nClasses &  FC/s1 & $512 \tiny{\times}$ nClasses &\\ \hline
 			  -&SoftMax/s1 & Classifier &  SoftMax/s1 & Classifier& \\ \hline
 			\end{tabular}
 	\end{center}
 	
 	\caption{Base classifier $C_O$ architectures. ``C'' is a conv. layer and ``FC'' a dense layer; ``s2'' denotes stride a stride of 2, while ``MP'' denotes a 2$\times$ 2MaxPool layer; For ResNet we add to the output $B$ of each block, the output of a $C/s1$ layer of shape $1\tiny{\times} 1 \tiny{\times} B^i_d\tiny{\times} B^o_d$ (with $B^i_d$/$B^o_d$ being the in-/output feature channels of the block); A BatchNorm and ReLU layer follows each ``C'' layer. The last column indicates layer (names) $L$ used for computing explanations.}  \label{tab:arch} 
 \end{table}

\noindent\emph{Explanations:} The explanations computed using the non-reflective network $C_O$ are based on existing methods, specifically an adapted version of GradCAM\cite{selvaraju2017grad}. Typically, attribution-based explanations (like GradCAM) aim to highlight relevant parts of the input creating an ``attention map'' with the same spatial extent as the input. However, our focus is not on attributing to the input itself, but rather on identifying the more semantically meaningful features at intermediate layers that have influenced a decision. To do this, we modify GradCAM to aggregate the channels of an intermediate layer rather than the input, resulting in multiple channels, each being a sum of a few other channels. This allows us to maintain spatial information while still simplifying the explanations through aggregation.

More formally, we obtain a class-discriminative, location-sensitive explanation $R^{c} \in \mathbf{R}^{u\times v \times d}$ of width $u$, height $v$ and depth $d$ for class $c$, where $R^{c}_i$ denotes the $i$-th channel. We explain the convolutional layer $L$ with $K$ feature maps $A^k \in \mathbf{R}^{u\times v}$, with each element indexed by $i,j$. That is, $A^k_{i,j}$ refers to the position $(i,j)$ of the feature map $A^k$, and the score of a class $y^c$ (before the softmax). \\
Explainability methods should allow understanding ``hypothetical'' outcomes, i.e., outcomes other than the actual predictions. This condition is fulfilled for commonly used methods that propagate information backward, e.g., using gradients\cite{selvaraju2017grad} or a relevance measure\cite{bach2015pixel}. Decoder-based and GAN-based methods such as those \cite{sch22exp,ngu16} that rely on forward pass information are less adequate. 
We focus on GradCAM\cite{selvaraju2017grad}.
GradCAM computes gradients concerning the score class $y^c$, i.e., the activation of that neuron. GradCAM performs global average pooling over the spatial dimensions height and width with indexes $i,j$ to compute neuron importance weights:
\begin{align} \alpha_k^c:= \frac{1}{Z} \sum_{i,j} \frac{\partial y^c}{\partial A^k_{i,j}} \end{align}
where $Z$ is a normalization constant. The weight can be interpreted as a partial linearization of the network downstream from activation maps $A$. It represents the importance of feature map $k$ for target class $c$. GradCAM aggregates all $K$ feature maps using their corresponding weights into a heatmap of depth 1 followed by a ReLU activation. In contrast, we aggregate by summing only $\frac{K}{d}$ feature maps rather than all $K$ (we assume $\frac{K}{d}$ yields an integer), and we do not employ a ReLU layer. That is, we get:
\begin{align}R^{c}_i = \sum_{k=\frac{K}{d}\cdot i}^{\frac{K}{d}\cdot (i+1)} \alpha_k^cA^k \end{align} 
For GradCAM, explanations of intermediate layers of smaller spatial dimensions are up-sampled to obtain attribution maps of the inputs. We reason in terms of the abstraction (and dimensions) provided by the explained layer and not of the original input.

\medskip
\noindent\emph{Reflective Network:} The reflective network $C_R$ has the same architecture as the classifier $C_O$ with a few extra layers to accommodate the explanations (Network in Figure \ref{fig:arch} including orange-colored layers). That is, the explanation is first processed using two sequential convolutional layers, each followed by batch normalization and a ReLU layer. The outcome of the second one is appended to the layer $L$ used to compute the explanations so that spatial dimensions are aligned. This reflective network $C_R$ can be trained with the same setup as the original classifier $C_O$, except that the input consists of the sample $X$ and the explanation $R^c$. For each labeled sample $(X,y)$, where $y$ is the ground truth for $X$, we choose the explained class ($Expl.Class$) randomly from the available explanations $Expl.Class_{Train}(X)$ in each iteration. However, the set $Expl.Class_{Train}(X)$ being a subset of explanations of all classes is static. That is, the explanations within $Expl.Class_{Train}(X)$ are pre-computed before training of the reflective network using the original classifier $C_O$. That is, explanations remain fixed throughout training. In particular, we consider the option that $Expl.Class_{Train}(X)$ contains an explanation of a randomly chosen class. In this case, we choose a class uniformly and independently at random among all possible classes for each set of $Expl.Class_{Train}(X)$ for each sample $X$ and maintain the same class for each sample throughout training.\footnote{Keeping explanations fixed throughout training is done to reduce computation time.}  For operation, i.e., testing, we consistently use the explanation of the prediction $y_p=C_O(X)$ of the non-reflective classifier, i.e., $Expl.Class_{Test}(X_0)=y_p$. In our empirical analysis, we also investigate other classes, e.g., using the ground truth $y_0$. However, in a real-world setting, the ground truth is usually not available at test time.

\subsection{Options for Architecture and Training} \label{sec:opt}
We describe design options to alter the base architecture shown in Figure \ref{fig:arch}. The actual architecture of the underlying classifier can be a VGG, ResNet, or any other neural network architecture.

\begin{itemize}
    \item Layer $L$: The layer we use for computing the explanation. That is, we backpropagate from the score of a class $c$, typically the last layer before the SoftMax, i.e., its logits layer, up to layer $L$ to obtain explanation $R^c$. Locations of layers $Middle$ and $High$ are indicated in Table \ref{fig:arch} for both architectures.
    \item Depth of the explanation $d$: The amount of information that an explanation contains, i.e., whether it is just a heatmap (one channel, $d=1$) or whether it contains multiple channels, i.e., $d>1$.
    \item Explained classes (= predictions) used for training $Expl.Class_{Train}$: We let $Expl.Class_{Train} \subseteq$ $\{correct, predicted, random\}$ for a dataset with $n$ classes. An explanation might be ``correct'', i.e., the one obtained from the correct output class. The predicted class of the network is denoted by $predicted$. ``Random'' refers to an explanation from a randomly chosen class of all classes $[1,n]$. That is, before training begins, if $random \in Expl.Class_{Train}$ we choose uniformly and independently a random class $y_r \in [1,n]$ for each sample $X$ and use this explanation for $y_r$. That is, we obtain a set of explanations $Expl.Class_{Train}(X)$ for each sample $X$. During training, the set $Expl.Class_{Train}(X)$ remains static. But at each epoch, sample $X$ is used for training  together with a random explanation of the set $Expl.Class_{Train}(X)$. The set can contain one or multiple explanations and even duplicates.    
    \item Explanation Source ($Expl.Source$): We consider three options: (i) Self: Fine-Tuning the same network used to obtain the explanations, resulting in self-reflection; all weights may be changed during fine-tuning; (ii) Other: A reflective network is trained from scratch using random initialization and explanations from another network; (iii) Noise: Explanations are random noise, with each value chosen uniformly at random in each iteration. The classifier should ignore these explanations, but it may still perform better due to retraining. We want to exclude the possibility that our improvements are only due to the effect of multiple resets of learning rates, or cyclic learning rates \cite{smi17}, which are known to improve performance. 
\end{itemize}

\section{Evaluation}
We conduct both qualitative and quantitative evaluations. For the quantitative evaluation, we investigate the impact of the architecture and training options on the performance of the reflective network (Section \ref{sec:opt}). We use visualization of explanations for the qualitative evaluation and use a recent explainability technique to understand the behavior of reflective networks.

\subsection{Setup and Analysis} \label{sec:set}
In our experiments, we used PyTorch 1.11.0 and Python 3.9 on an Ubuntu machine with an NVIDIA RTX 2080 Ti. We used a default setup, with some parameters varied for individual experiments. Stochastic Gradient Descent with momentum 0.9 with batchsize 128 was used for training. Overfitting was generally not a major concern, so we only reported test performance after training.The initial learning rate of 0.1 was decreased by 0.1 at epochs 70 and 120 of the 150 epochs for training the original classifier $C_O$, and half as many epochs for retraining $C_R$. We also employed weight decay with a parameter of 0.0005. Unless otherwise specified, we used a VGG and ResNet variant (see Table \ref{tab:arch}) and the following settings (Section \ref{sec:opt}):$d=32$, $Expl.Class_{Train}=\{correct,predicted,random\}$, $Exp_{Source}=Self$. For VGG, we used the $L=Middle$ layer as indicated in the last column in Table \ref{tab:arch}, and for ResNet, $L=High$ was the second conv layer of the block. We used CIFAR-10/100, SVHN\cite{net11}, FashionMNIST\cite{xia17} (scaled to 32x32), and TinyImageNet\cite{stan18} without data augmentation. Since the TinyImageNet architecture has a larger spatial extent (64x64), more classes, and more data, we added a block to ResNet and two layers to VGG for downsampling; we also scaled the number of neurons of all layers by a factor of 1.5 and used $d=64$ to account for the growth in channels also when learning from explanations. Predefined splits into training and test data were employed. We trained 7 networks for each configuration. We report the average accuracy and standard deviation (Table \ref{tab:arch}). Code is available at \url{https://github.com/JohnTailor/Reflective-Net-Learning-from-Explanations}.

\subsection{Quantitative}
\begin{table}
    \setlength\tabcolsep{5pt}
	\centering
	\small
	\begin{tabular}{|c|c|c|c|c||c|	} \hline 
& \multicolumn{2}{|c|}{Cifar-100}& \multicolumn{2}{|c|}{Cifar-10}&\multirow{3}{*}{Average    }\\ \cline{2-5}
& \multicolumn{1}{|c|}{ ResNet}& \multicolumn{1}{|c|}{VGG}& \multicolumn{1}{|c|}{ ResNet}& \multicolumn{1}{|c|}{VGG} &\\ \hline
Baseline (Non-Reflective) &\textbf{59.0}\text{\tiny{$\pm$0.4}}&\textbf{57.3}\text{\tiny{$\pm$0.3}}&\textbf{86.8}\text{\tiny{$\pm$0.2}}&\textbf{84.7}\text{\tiny{$\pm$0.3}}& \\ \hline
\noalign{\vskip 2mm} \hline    \emph{Depth $d$} & \multicolumn{5}{|c|}{}\\ \hline 1&0.4\text{\tiny{$\pm$0.5}}&0.3\text{\tiny{$\pm$0.6}}&-1.4\text{\tiny{$\pm$0.7}}&-0.5\text{\tiny{$\pm$0.6}}&-0.29\\ \hline 
4&0.9\text{\tiny{$\pm$0.7}}&0.4\text{\tiny{$\pm$0.5}}&-0.1\text{\tiny{$\pm$0.5}}&-0.0\text{\tiny{$\pm$0.4}}&0.3\\ \hline 
32&2.2\text{\tiny{$\pm$0.6}}&1.0\text{\tiny{$\pm$0.4}}&0.2\text{\tiny{$\pm$0.4}}&\textbf{0.4\text{\tiny{$\pm$0.3}}}&0.92\\ \hline 
128&\textbf{2.6\text{\tiny{$\pm$0.5}}}&\textbf{1.4\text{\tiny{$\pm$0.4}}}&\textbf{0.3\text{\tiny{$\pm$0.3}}}&-1.5\text{\tiny{$\pm$0.3}}&0.71\\ \hline
\noalign{\vskip 2mm} \hline    \emph{$Expl.Class_{Train}$} & \multicolumn{5}{|c|}{}\\ \hline correct&-1.0\text{\tiny{$\pm$0.1}}&-1.3\text{\tiny{$\pm$0.4}}&-0.9\text{\tiny{$\pm$0.1}}&-0.7\text{\tiny{$\pm$0.2}}&-0.96\\ \hline 
predicted&-1.0\text{\tiny{$\pm$0.1}}&-1.1\text{\tiny{$\pm$0.6}}&-0.9\text{\tiny{$\pm$0.1}}&-0.6\text{\tiny{$\pm$0.1}}&-0.89\\ \hline 
random&1.5\text{\tiny{$\pm$0.5}}&0.3\text{\tiny{$\pm$1.0}}&-0.7\text{\tiny{$\pm$0.2}}&-4.0\text{\tiny{$\pm$1.0}}&-0.73\\ \hline 
correct, predicted&-0.9\text{\tiny{$\pm$0.2}}&-1.3\text{\tiny{$\pm$0.6}}&-0.9\text{\tiny{$\pm$0.1}}&-0.6\text{\tiny{$\pm$0.2}}&-0.96\\ \hline 
correct, random&\textbf{2.7\text{\tiny{$\pm$0.5}}}&0.9\text{\tiny{$\pm$0.6}}&\textbf{0.3\text{\tiny{$\pm$0.2}}}&-0.0\text{\tiny{$\pm$0.2}}&0.98\\ \hline 
correct, predicted, random&2.2\text{\tiny{$\pm$0.6}}&\textbf{1.0\text{\tiny{$\pm$0.4}}}&0.2\text{\tiny{$\pm$0.4}}&\textbf{0.4\text{\tiny{$\pm$0.3}}}&0.92\\ \hline
\noalign{\vskip 2mm} \hline    \emph{Expl.Source} & \multicolumn{5}{|c|}{}\\ \hline Self&\textbf{2.2\text{\tiny{$\pm$0.6}}}&1.0\text{\tiny{$\pm$0.4}}&0.2\text{\tiny{$\pm$0.4}}&\textbf{0.4\text{\tiny{$\pm$0.3}}}&0.92\\ \hline 
Other&1.9\text{\tiny{$\pm$0.5}}&\textbf{1.3\text{\tiny{$\pm$0.4}}}&\textbf{0.9\text{\tiny{$\pm$0.3}}}&-0.4\text{\tiny{$\pm$0.8}}&0.92\\ \hline 
Noise&1.4\text{\tiny{$\pm$0.6}}&0.9\text{\tiny{$\pm$0.6}}&-0.6\text{\tiny{$\pm$0.1}}&-0.1\text{\tiny{$\pm$0.2}}&0.4\\ \hline 
\noalign{\vskip 2mm} \hline    \emph{Layer $L$} & \multicolumn{5}{|c|}{}\\ \hline Middle&1.2\text{\tiny{$\pm$0.7}}&0.3\text{\tiny{$\pm$0.6}}&-0.4\text{\tiny{$\pm$0.4}}&0.4\text{\tiny{$\pm$0.3}}&0.36\\ \hline 
Middle, High&2.2\text{\tiny{$\pm$0.5}}&\textbf{1.1\text{\tiny{$\pm$0.4}}}&\textbf{0.6\text{\tiny{$\pm$0.4}}}&\textbf{0.4\text{\tiny{$\pm$0.3}}}&1.06\\ \hline 
High&\textbf{2.3\text{\tiny{$\pm$0.3}}}&1.1\text{\tiny{$\pm$0.2}}&0.2\text{\tiny{$\pm$0.3}}&0.4\text{\tiny{$\pm$0.3}}&0.99\\ \hline

\end{tabular}
\vspace{6pt}
	\caption{Empirical analysis using explanations of predictions $Expl.Class_{Test}=predicted$. Numbers indicate absolute differences in accuracy to a non-reflective network used as baseline stated in the top row (no data augmentation was used). Larger numbers are better. Average is the mean of the differences of all four settings.}\label{tab:abl}
\end{table}

\begin{table}
    \setlength\tabcolsep{2.5pt}
	\centering
	\small
	\begin{tabular}{|c| c|c|c| c|c|c| 	} \hline 
& \multicolumn{3}{|c|}{Cifar-100/ ResNet}& \multicolumn{3}{|c|}{Cifar-100/ VGG}\\ \cline{2-7}
& \multicolumn{3}{|c|}{$Expl.Class_{Test}$}& \multicolumn{3}{|c|}{$Expl.Class_{Test}$}\\ \cline{2-7}
& predicted & correct& random& predicted & correct& random\\ \hline
Baseline (Non-Reflective) &\multicolumn{3}{|c|}{\textbf{59.0}\text{\tiny{$\pm$0.4}}}&\multicolumn{3}{|c|}{\textbf{57.3}\text{\tiny{$\pm$0.3}}} \\ \hline
\noalign{\vskip 2mm} \hline    \emph{Depth $d$} & \multicolumn{6}{|c|}{}\\ \hline 1&0.4\text{\tiny{$\pm$0.5}}&0.7\text{\tiny{$\pm$0.8}}&0.2\text{\tiny{$\pm$0.5}}&0.3\text{\tiny{$\pm$0.6}}&0.3\text{\tiny{$\pm$0.6}}&0.2\text{\tiny{$\pm$0.6}}\\ \hline 
4&0.9\text{\tiny{$\pm$0.7}}&1.7\text{\tiny{$\pm$1.0}}&\textbf{0.5\text{\tiny{$\pm$0.8}}}&0.4\text{\tiny{$\pm$0.5}}&0.4\text{\tiny{$\pm$0.6}}&\textbf{0.4\text{\tiny{$\pm$0.6}}}\\ \hline 
32&2.2\text{\tiny{$\pm$0.6}}&9.3\text{\tiny{$\pm$3.1}}&-0.4\text{\tiny{$\pm$0.7}}&1.0\text{\tiny{$\pm$0.4}}&2.2\text{\tiny{$\pm$0.7}}&0.4\text{\tiny{$\pm$0.4}}\\ \hline 
128&\textbf{2.6\text{\tiny{$\pm$0.5}}}&\textbf{15.5\text{\tiny{$\pm$0.6}}}&-0.7\text{\tiny{$\pm$0.7}}&\textbf{1.4\text{\tiny{$\pm$0.4}}}&\textbf{5.4\text{\tiny{$\pm$0.6}}}&-0.6\text{\tiny{$\pm$0.6}}\\ \hline
\noalign{\vskip 2mm} \hline    \emph{$Expl.Class_{Train}$} & \multicolumn{6}{|c|}{}\\ \hline correct&-1.0\text{\tiny{$\pm$0.1}}&24.4\text{\tiny{$\pm$0.5}}&-25.8\text{\tiny{$\pm$1.7}}&-1.3\text{\tiny{$\pm$0.4}}&4.7\text{\tiny{$\pm$0.2}}&-5.7\text{\tiny{$\pm$0.9}}\\ \hline 
predicted&-1.0\text{\tiny{$\pm$0.1}}&24.6\text{\tiny{$\pm$0.2}}&-25.1\text{\tiny{$\pm$1.5}}&-1.1\text{\tiny{$\pm$0.6}}&\textbf{5.1\text{\tiny{$\pm$0.4}}}&-5.8\text{\tiny{$\pm$0.7}}\\ \hline 
random&1.5\text{\tiny{$\pm$0.5}}&1.5\text{\tiny{$\pm$0.5}}&\textbf{1.5\text{\tiny{$\pm$0.5}}}&0.3\text{\tiny{$\pm$1.0}}&0.4\text{\tiny{$\pm$1.0}}&0.4\text{\tiny{$\pm$0.8}}\\ \hline 
correct, predicted&-0.9\text{\tiny{$\pm$0.2}}&\textbf{24.6\text{\tiny{$\pm$0.5}}}&-25.4\text{\tiny{$\pm$1.0}}&-1.3\text{\tiny{$\pm$0.6}}&4.6\text{\tiny{$\pm$0.3}}&-5.9\text{\tiny{$\pm$0.7}}\\ \hline 
correct, random&\textbf{2.7\text{\tiny{$\pm$0.5}}}&8.4\text{\tiny{$\pm$0.6}}&0.8\text{\tiny{$\pm$0.8}}&0.9\text{\tiny{$\pm$0.6}}&1.8\text{\tiny{$\pm$0.4}}&\textbf{0.6\text{\tiny{$\pm$0.6}}}\\ \hline 
correct, predicted, random&2.2\text{\tiny{$\pm$0.6}}&9.3\text{\tiny{$\pm$3.1}}&-0.4\text{\tiny{$\pm$0.7}}&\textbf{1.0\text{\tiny{$\pm$0.4}}}&2.2\text{\tiny{$\pm$0.7}}&0.4\text{\tiny{$\pm$0.4}}\\ \hline
\noalign{\vskip 2mm} \hline    \emph{Expl.Source} & \multicolumn{6}{|c|}{}\\ \hline Self&\textbf{2.2\text{\tiny{$\pm$0.6}}}&\textbf{9.3\text{\tiny{$\pm$3.1}}}&-0.4\text{\tiny{$\pm$0.7}}&1.0\text{\tiny{$\pm$0.4}}&2.2\text{\tiny{$\pm$0.7}}&0.4\text{\tiny{$\pm$0.4}}\\ \hline 
Other&1.9\text{\tiny{$\pm$0.5}}&7.7\text{\tiny{$\pm$0.9}}&-0.7\text{\tiny{$\pm$0.8}}&\textbf{1.3\text{\tiny{$\pm$0.4}}}&\textbf{2.5\text{\tiny{$\pm$0.4}}}&0.6\text{\tiny{$\pm$0.5}}\\ \hline 
Noise&1.4\text{\tiny{$\pm$0.6}}&1.4\text{\tiny{$\pm$0.6}}&\textbf{1.4\text{\tiny{$\pm$0.7}}}&0.9\text{\tiny{$\pm$0.6}}&0.9\text{\tiny{$\pm$0.6}}&\textbf{0.9\text{\tiny{$\pm$0.6}}}\\ \hline 
\noalign{\vskip 2mm} \hline    \emph{Layer $L$} & \multicolumn{6}{|c|}{}\\ \hline Middle &1.2\text{\tiny{$\pm$0.7}}&1.4\text{\tiny{$\pm$0.7}}&\textbf{1.0\text{\tiny{$\pm$0.7}}}&0.3\text{\tiny{$\pm$0.6}}&0.5\text{\tiny{$\pm$0.6}}&0.2\text{\tiny{$\pm$0.5}}\\ \hline 
Middle, High&2.2\text{\tiny{$\pm$0.5}}&\textbf{10.6\text{\tiny{$\pm$0.6}}}&-0.7\text{\tiny{$\pm$0.6}}&\textbf{1.1\text{\tiny{$\pm$0.4}}}&\textbf{2.6\text{\tiny{$\pm$0.5}}}&0.1\text{\tiny{$\pm$0.6}}\\ \hline 
High&\textbf{2.3\text{\tiny{$\pm$0.3}}}&10.5\text{\tiny{$\pm$0.4}}&-0.6\text{\tiny{$\pm$0.4}}&1.1\text{\tiny{$\pm$0.2}}&2.4\text{\tiny{$\pm$0.2}}&\textbf{0.4\text{\tiny{$\pm$0.3}}}\\ \hline

\noalign{\vskip 6mm} \hline
 & \multicolumn{3}{|c|}{ResNet} & \multicolumn{3}{|c|}{VGG}\\ \cline{2-7} 
& \multicolumn{3}{|c|}{$Expl.Class_{Test}$}& \multicolumn{3}{|c|}{$Expl.Class_{Test}$}\\ \hline
\emph{Datasets} & predicted & correct& random& predicted & correct& random\\ \hline

\noalign{\vskip 1mm} \hline
FashionMNIST & \multicolumn{6}{|c|}{}\\ \hline
Baseline (Non-Reflective) &\multicolumn{3}{|c|}{\textbf{93.0}\text{\tiny{$\pm$0.1}}}&\multicolumn{3}{|c|}{\textbf{92.8}\text{\tiny{$\pm$0.2}}} \\ \hline
Reflective-Net    &\textbf{0.6\text{\tiny{$\pm$0.1}}}&\textbf{1.9\text{\tiny{$\pm$0.2}}}&\textbf{0.1\text{\tiny{$\pm$0.1}}}&\textbf{0.2\text{\tiny{$\pm$0.0}}}&\textbf{0.9\text{\tiny{$\pm$0.1}}}&\textbf{-0.1\text{\tiny{$\pm$0.3}}}\\ \hline

\noalign{\vskip 1mm} \hline SVHN & \multicolumn{6}{|c|}{}\\ \hline
Baseline (Non-Reflective) &\multicolumn{3}{|c|}{\textbf{95.1}\text{\tiny{$\pm$0.3}}}&\multicolumn{3}{|c|}{\textbf{94.7}\text{\tiny{$\pm$0.3}}} \\ \hline
Reflective-Net&\textbf{0.2\text{\tiny{$\pm$0.2}}}&\textbf{1.7\text{\tiny{$\pm$0.3}}}&\textbf{-0.7\text{\tiny{$\pm$0.4}}}&\textbf{0.1\text{\tiny{$\pm$0.1}}}&\textbf{0.5\text{\tiny{$\pm$0.0}}}&\textbf{-0.2\text{\tiny{$\pm$0.3}}}\\ \hline

\noalign{\vskip 1mm} \hline Tiny-ImageNet & \multicolumn{6}{|c|}{}\\ \hline
Baseline (Non-Reflective) &\multicolumn{3}{|c|}{\textbf{45.6}\text{\tiny{$\pm$0.2}}}&\multicolumn{3}{|c|}{\textbf{43.3}\text{\tiny{$\pm$0.6}}} \\ \hline
    Reflective-Net&\textbf{2.3\text{\tiny{$\pm$0.2}}}&\textbf{8.6\text{\tiny{$\pm$0.6}}}&\textbf{-0.4\text{\tiny{$\pm$0.5}}}&\textbf{0.4\text{\tiny{$\pm$0.8}}}&\textbf{-0.2\text{\tiny{$\pm$1.1}}}&\textbf{-0.6\text{\tiny{$\pm$1.1}}}\\ \hline    
\end{tabular}

\vspace{6pt}
	\caption{Empirical analysis showing only Cifar-100. Numbers indicate differences in accuracy in percent to a non-reflective network stated as baseline in the top row (no data augmentation was used). Larger is better. Bold is best in the column of a hyperparameter. }\label{tab:abl2}
\end{table}

\begin{figure}[ht]
 \centering{  \includegraphics[width=0.7\linewidth]{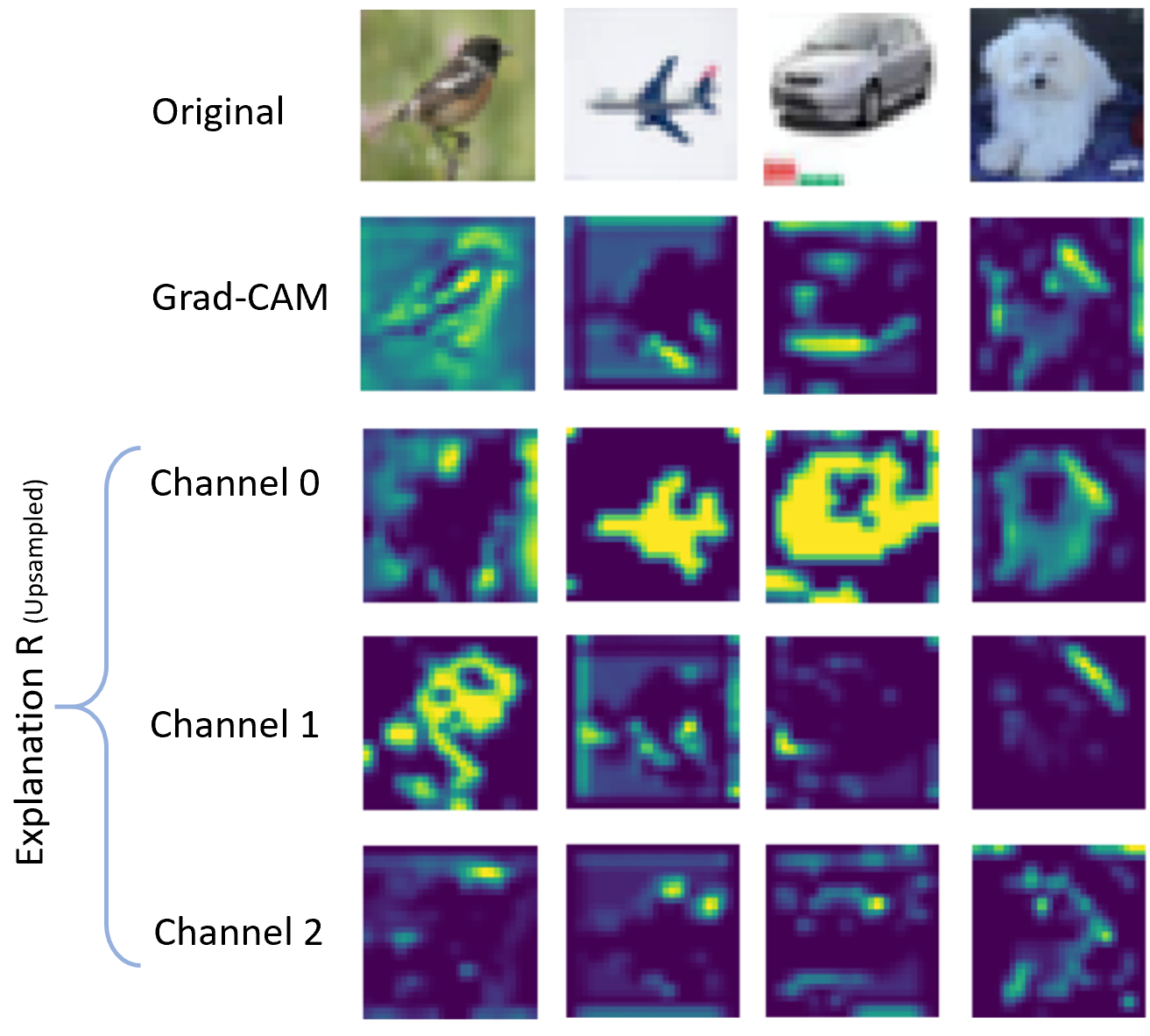}}
 \caption{Comparison of explanations using GradCAM and upsampled explanations of VGG on CIFAR-10 after the second conv layer ($L=Low$ in Table \ref{fig:arch}) } \label{fig:exps}
\end{figure}
We perform first an empirical analysis using two architectures and two datasets. We then evaluate  on additional datasets.

\medskip
\noindent \textbf{Empirical Analysis}: The results of varying design parameters can be seen in Tables \ref{tab:abl} and \ref{tab:abl2}. The main findings are: it is important to train with multiple explanations per input, explanations based on upper layers are more effective, and using explanations in general is beneficial, as demonstrated by the "sanity" checks (such as using random explanations or extensive compression ($d=1$)).
Table \ref{tab:abl} focuses on the most common scenario, where the predicted class is used ($Expl.Class_{Test}=predicted$) to obtain an explanation during test time, regardless of whether it is correct or not. Table \ref{tab:abl2} includes a single dataset and additional results for explanations of random and correct classes at test time, i.e., $Expl.Class_{Test}=random$ and $Expl.Class_{Test}=correct$. 

\noindent\emph{$Expl.Class_{Train}$:} Training with only a single explanation per sample, i.e., without reflecting on multiple options, only yields consistent gains for random explanations, i.e., $Expl.Class_{Train}=random$. This gain can be attributed to retraining of the network, as the reflective network can be considered equivalent to a classifier trained twice with a decaying learning rate. Such cyclic learning rates are known to be beneficial \cite{smi17}. When using only predicted explanations, i.e., $Expl.Class_{Train}=predicted$, during training, the classifier tends to rely on the explanations rather than the actual inputs and is unable to improve upon the original classifier except when the ground truth is used at test time. However, if explanations for a random class are used, the classifier performs poorly because it predicts the randomly chosen class of these explanations regardless of the actual input.

On the other hand, choosing randomly between an explanation for a random (likely incorrect) class and the predicted or correct class leads to significant performance gains across all architectures and datasets. It does not matter whether the explained classes, i.e., $Expl.Class_{Train}$, contain both the predicted and correct classes or just the correct and random classes. Note that in the former case, the correct explanation is often contained twice within a set $Expl.Class_{Train}(X)$ for a sample $X$, but this imbalance has little impact. This suggests that learning by considering multiple options is necessary and that "what-if" type of reasoning during training, even with explanations based on classes different from the correct or predicted ones, is very helpful. Additionally, it is important to ensure that the network does not rely solely on explanations.

\medskip
\noindent\emph{Depth $d$:}  Using a heatmap ($d=1$) as an explanation does not yield much improvement. This suggests that spatial information is not particularly valuable, as heatmaps only highlight locations without providing semantic information. On the other hand, using more detailed explanations that provide information on features is beneficial. In general, using more detailed explanations (larger $d$) is helpful, with the exception of VGG on Cifar-10. A smaller $d$ results in explanations that contain less information and are less discriminative, while larger $d$ provides class-specific information. For example, heatmaps ($d=1$) highlight the center of an image regardless of the class, while larger $d$ explanations provide information relevant to the class. When $d$ is at its maximum, i.e., no aggregation occurs, each value is the product of the feature activation and the gradient. For features that are not relevant to the object, this product will be zero due to the activation being zero. However, using overly fine-grained explanations carries the risk of overfitting, as the training data may contain few samples that exhibit specific activation and explanation patterns that are not generalizable to the test data. Aggregation can reduce the risk of heavily relying on such patterns.\\

\medskip
\noindent\emph{$Expl.Source$ and layer $L$:} The layer used to obtain the explanation has some influence on the results. Using explanations based on multiple layers can be advantageous, and upper layers tend to perform better overall. This might be because it is easier to identify a potential mismatch between activations from the forward pass and the explanation when the features are more specialized towards a specific class. As for the explanation source, there are varying differences in accuracy between training a network from scratch or fine-tuning a pre-trained network. Training networks from scratch ($Expl.Source=Other$) allows the network to learn more freely from the explanations. Self-training benefits from the cyclic training effect mentioned earlier \cite{smi17}, and using noise for explanations also leads to improvements due to this effect.  \\

\begin{figure}[ht]
 \centering{  \includegraphics[width=\linewidth]{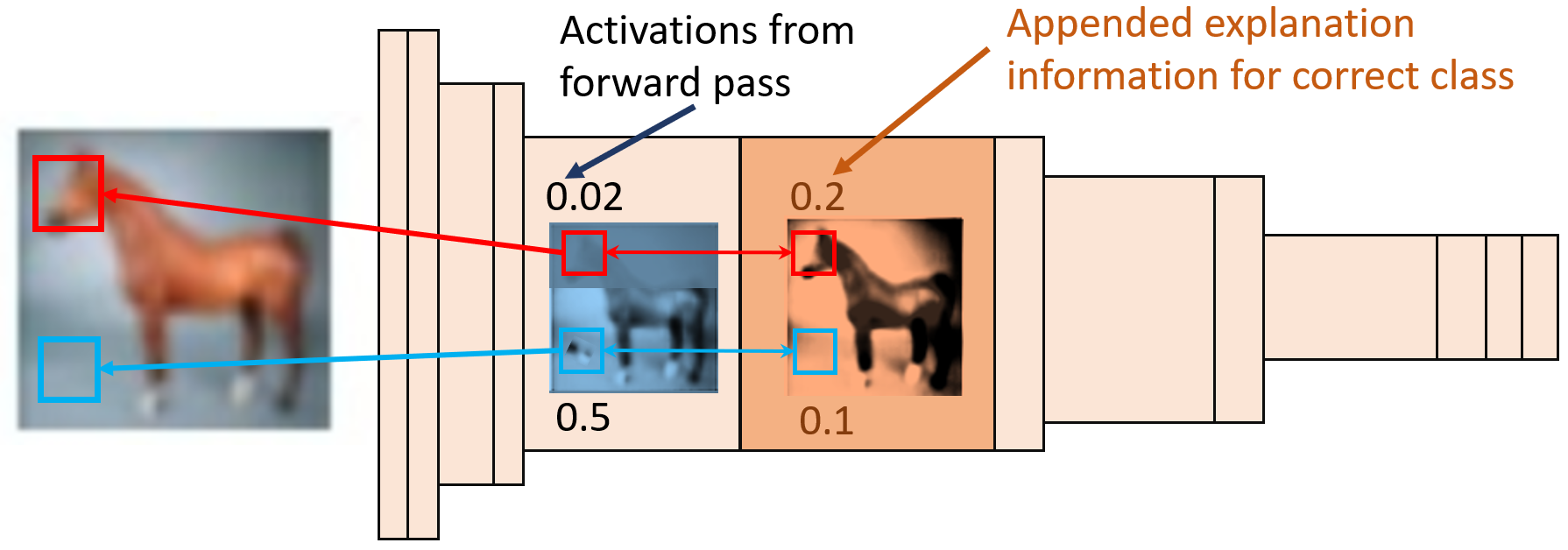}}
 \caption{Explanation for correct class for an incorrect prediction. Blue shows an irrelevant feature for the correct class that still shows strong activations for the forward pass. Red shows a feature that should show strong activation according to the explanation but does not. Thus, an explanation of the correct class might provide more adequate information for classification.} \label{fig:fix}
\end{figure}

\noindent\emph{$Expl.Class_{Test}$:} As shown in Table \ref{tab:abl2}, using the explanation for the correct class during testing, i.e., $Expl.Class_{Test}=correct$, generally leads to significant improvements, often exceeding 15\%, compared to not using explanations. This demonstrates the importance of understanding the reasons or features at a lower level that contribute to the final decision. However, it is worth noting that at the layer $L$ where the explanation is obtained, the classifier may already be heading in the wrong direction, with some features deemed relevant (having positive activation) that are actually irrelevant for the correct class, while others deemed irrelevant but are actually highly relevant. This could be due to sensitivity to noise, for example. If the explanation for the correct class is used, the classifier can overcome errors caused by imprecise recognition in lower layers and focus on the features that should exhibit strong activations for the correct class prediction but are not doing so. For example, if the classifier predicts ``cat" but should predict ``horse," and the explanation highlights a relevant feature related to a horse's head that should have strong activation but only has weak activation during the forward pass due to noise or inadequate representation in lower layers, the explanation can compensate for this absence and help the network rely on it to some extent, as shown in Figure \ref{fig:fix}.\\
Using explanations for a random class during testing, i.e., $Expl.Class_{Test}=random$, leads to lower accuracy than the non-reflective network, i.e., the classifier $C_O$ that does not use explanations. This is expected because these explanations often point to a wrong class unrelated to the correct class or prediction. The impact is greatest if only the correct or predicted class is used during training, as the network learns to trust the explanations and believes they are likely correct. However, if this is not the case during testing due to the use of random explanations, the network relies on this incorrect information and performs poorly.

\medskip
\noindent\textbf{Various Datasets}:
Table \ref{tab:abl2} presents the results for the TinyImageNet, FashionMNIST, and SVHN datasets. For FashionMNIST, the base classifier without explanations already performs very well with accuracies above 90\%, and training with reflections only leads to small improvements. On the other hand, for TinyImageNet, explanations based on predictions are clearly beneficial. Additionally, the results for CIFAR-10 and CIFAR-100 (Table \ref{tab:abl2}) suggest that reflections are less effective on simple datasets with a few classes and easy classification, compared to more complex datasets where the baseline classifiers do not perform exceptionally well. In these cases, the cyclic training method appears to be more effective.

\subsection{Qualitative Evaluation}
\noindent\emph{Explanations:} Explanations from GradCAM and our explanations show similarities. This can be seen in Figure \ref{fig:exps}. All images are normalized to span the entire range from 0 to 1. Note, we only show the first three of 16 channels, we use the second layer $L=Low$ (Table \ref{fig:arch}) and explanations are upsampled to have the same dimensions as GradCAM. Mathematically, GradCAM explanations are the sum of the (channel-wise) explanations, so overlap is expected. The channel-wise explanations seem to emphasize some areas much more than GradCAM does, e.g., the explanation of channel 0 for the car seems to indicate that almost the entire car is highly relevant. However, these differences are mostly due to normalization. They might also be a consequence of cancellation, e.g., in GradCAM, two feature maps might be the inverse of each other and, therefore, their additions lead to cancellation. Compared to using a single channel as in GradCAM, the channel-wise explanation give a more nuanced view of what areas and features impact the prediction.\\

\medskip

\begin{figure*}[ht]
 \centering{  \includegraphics[width=0.95\linewidth]{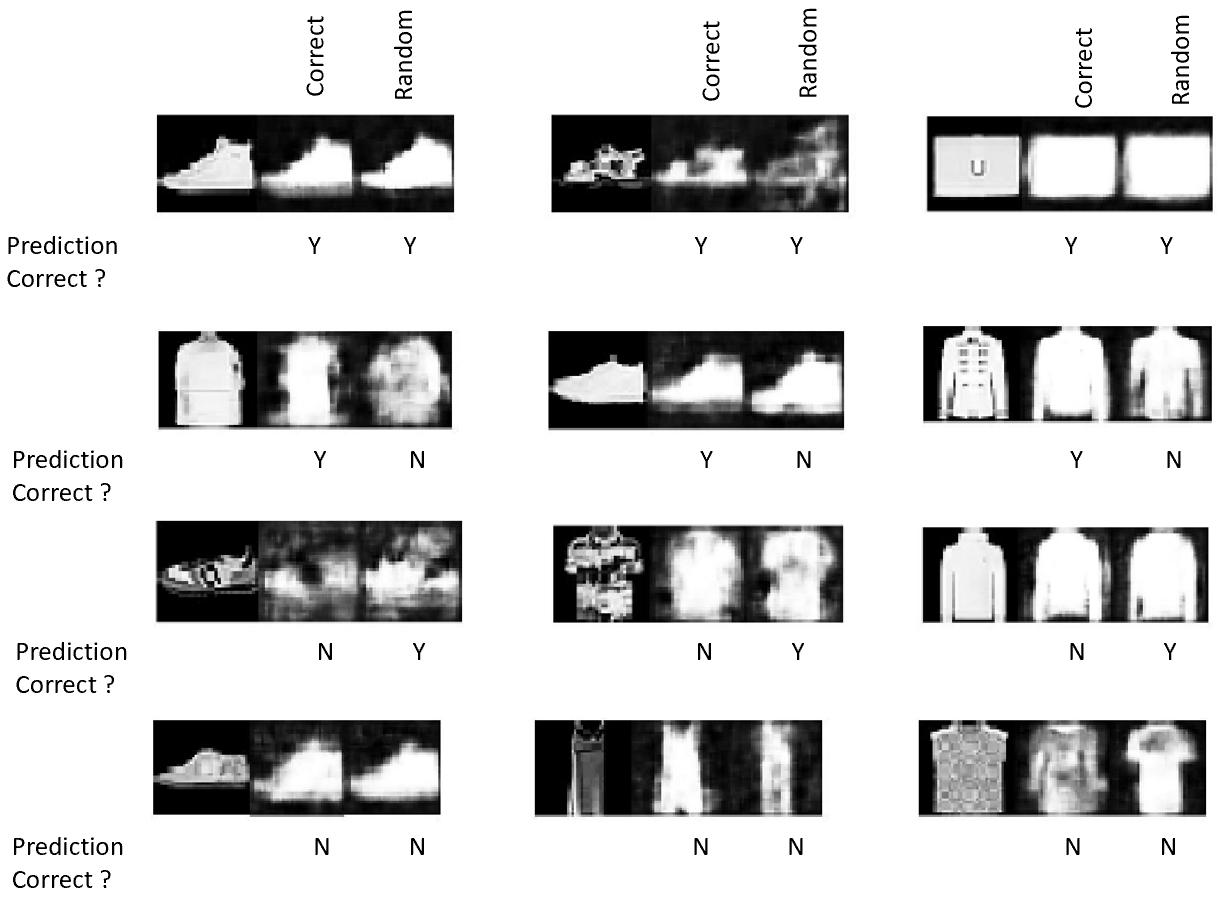}}
 \caption{Understanding reflective networks based on correct and random explanations at test time using a recent explainability technique called ClaDec\cite{sch22exp}. Random explanations at test time tend to mislead the classifier. } \label{fig:clad}
\end{figure*}

\noindent\emph{Reconstructions from Activations:}
In order to investigate the influence of explanations on the output, we applied a recently developed explainability technique called ClaDec\cite{sch22exp}. This method trains a decoder from layer activations to reconstruct the original inputs to the classifier. The reconstructions highlight what concepts of the input maintain throughout the layers, i.e., which of them are part of the current activations. Unlike techniques like GradCAM, which only provide attribution, reconstructions provide more detailed information about what information the network is using and how it encodes it. We used the open-source implementation of ClaDec and a reflective network trained on the FashionMNIST dataset using our default setup. ClaDec was trained to reconstruct input images\footnote{We only reconstructed input images. It is also possible to train to reconstruct explanations, which are also part of the input to the reflective network. However, we are primarily interested in understanding how activations relate to input images.} based on activations from the dense layer (the last layer before SoftMax) that were generated by feeding an input image and either the explanation for the correct class or a random class to the classifier.\\
Figure \ref{fig:clad} shows reconstructions in three columns: the ground truth (the original image to be classified), the reconstruction based on activations from the input and of the correct explanation, and, finally, a random explanation. We also indicate whether the prediction was correct. It can be observed that random explanations tend to mislead the network, resulting in inaccurate reconstructions. For example, in the first row, the second image is a mix of a sandal (correct class) and a boot (from a random explanation). Incorrect explanations can also lead to unusual activation patterns that are not accurately reconstructed into any meaningful object. For example, the first column in the second row shows a bag. The shape of the bag is atypical (in the dataset) and resembles a T-shirt to some degree. Using the correct explanation, the reconstruction is not perfect, but it is closer in shape to a typical bag in the data than the ground truth. For the random explanation, the reconstruction is very blurry and shows some weak elements of the wrong explanation (sweatshirt). In many cases, explanations have little impact -- for instance, both reconstructions in the second row, middle column, are similar, although the random one is for an incorrect class (boot) rather than the correct class (sneaker).\\

\medskip

\begin{figure*}
 \centering{  \includegraphics[width=0.95\linewidth]{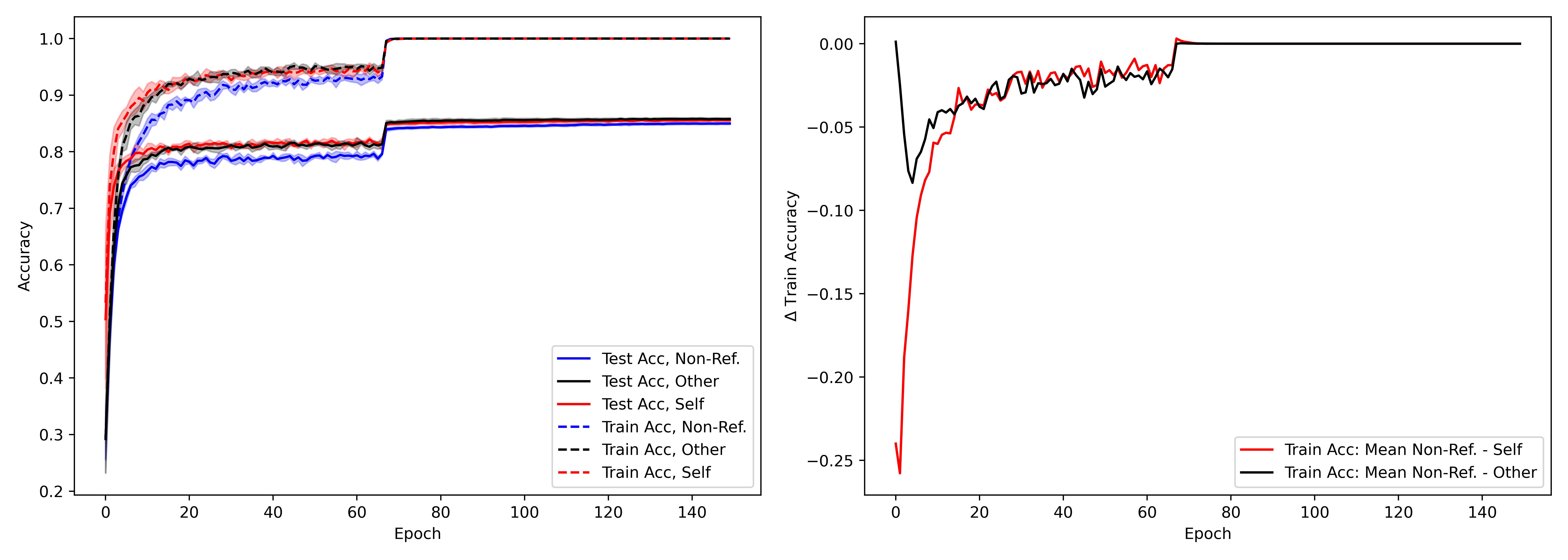}}
 \caption{Convergence of VGG on Cifar-10; left panel:learning curves; right panel: difference of training accuracy between non-reflective and reflective network w/o fine-tuning;} \label{fig:speed}
\end{figure*}

\noindent\emph{Training iterations:} The left panel in Figure \ref{fig:speed} illustrates the learning curves of the original classifier without explanation and the reflective network with and without fine-tuning (``self''-reflection and ``other''). Fine-tuning involves adjusting all weights and results in the quickest convergence. Interestingly, the reflective network without fine-tuning, where all parameters are randomly initialized, still converges faster than the non-reflective network. This can be best seen in the right panel of Figure \ref{fig:speed}. This can be observed in the right panel, which displays the differences in training accuracy between reflective and non-reflective networks. Initially, the differences are substantial, with reflective networks with fine-tuning showing more than 20\% improvement. This indicates that reflective networks learn faster initially.  Only when the learning rate is decayed, training accuracy of non-reflective and reflective networks reach 100\% although the gap in test accuracy remains in favor of reflective networks.
Overall, using a reflective process reduces the number of training iterations but may not necessarily lower computational costs, as explanations must still be calculated.

\section{Related Work}

\noindent\emph{Data Augmentation:} In data augmentation, the input data is typically modified or synthesized using techniques such as geometric transformations on images or GANs \cite{shor19,cub20}. Data augmentation can alter both the size and quality of the training data. Explanations, which can have spatial and depth information, can also be considered as input data and therefore can be augmented using similar techniques. In this work, we only augmented explanations by adding Gaussian noise. Adding an explanation of a fixed class, such as the predicted one, to the input does not alter the original input (image), but rather adds additional information generated by the explainability method. Using explanations from multiple classes and selecting one of them randomly during training could be considered a form of data augmentation that increases the quantity of training data. However, the choice of classes for which explanations are computed is restricted, as our experimental results show that the correct class should be included and using too many classes is not beneficial. In contrast, data augmentation on images typically does not have such constraints, and there is no requirement that the unmodified image must be part of the training data or that only a few combinations of augmentations give the best results. However, augmentations are often specific to the domain and may have limits on their strength, for example, rotating an image by 180 degrees could change its meaning, such as turning a ``3'' into an ``E''.

\medskip
\noindent\emph{Data enhancement:} Data enhancement involves adding extra data elements to an existing dataset. The trend of big data and the use of machine learning have increased the efforts to combine multiple data sources, leading to a growing interest in multi-modal learning \cite{roh19,bal18,bay21}. This is motivated by the fact that our world is multi-modal, meaning that humans can perceive information through multiple senses such as sight, sound, and smell. Using multiple modes of information, such as image data and its accompanying text, can improve prediction tasks. Our work is multi-modal in the sense that we use both the actual observations and self-generated data, comprising a visual mode for images and a thinking mode for reflection.

\medskip
\noindent\emph{Learning methods:} Backpropagation \cite{rum86} is a technique for adjusting the weights of a network iteratively to minimize the difference between the actual output and the desired output using gradient descent. Feedback is known to be important in the brain \cite{whi19,lil20}, as it can alter neural activity, which is not captured by backpropagation \cite{lil20}. Other models, such as hierarchical Bayesian inference \cite{lee03} or Helmholtz machines \cite{day95}, do take feedback into account. In our work, we rely on backpropagation and incorporate feedback from higher layers across networks by using explanations from a classifier as input to a reflective network. This is similar to the popular theory \cite{dan11} that suggests that a classifier first ``thinks" quickly (during a one-time, initial pass) and then ``thinks" slowly, through a slower process of explanation that involves computing the explanation and using it as input to the reflective network. The recirculation algorithm \cite{hin88,bio96} has been used for closed-loop networks and also involves two passes, with the first pass consisting of a visible vector circulating the loop and the recirculation consisting of an average of the visible vector and a reconstruction error. Its generalized form \cite{bio96} is considered to be more biologically plausible.
Explanation-based learning was first introduced in the late 1980s \cite{ell89}, but it differs from our research and the current understanding of XAI. Explanation-based learning is more similar to "one-shot learning," in which a single training sample is used to derive general rules that describe the behavior of the system. These rules are viewed as explanations in \cite{ell89}.    

\medskip
\noindent\emph{Attention with Gradients:} Attention mechanisms \cite{vasw17} have been widely applied to image recognition and localization tasks \cite{wang17,woo18,bel19,zha20}. These methods typically involve encoding attention through a mask \cite{wang17} that is multiplied with the feature map. Some approaches aim to cover more than just the most discriminative aspect of an object, which is often the focus of methods like GradCAM \cite{jia19,cho19,li2018,jet18}. For example, \cite{jia19} averaged attention maps from multiple training epochs, \cite{cho19} used attention-based dropout, \cite{li2018} incorporated a classification loss and an attention mining loss, and \cite{wan19} used multiple loss terms in addition to the classification loss, including attention separability and consistency loss terms. \cite{fuk19} introduced an attention-branch network that combines a response-based visual explanation model with an attention mechanism on a classifier branch, while \cite{els19} used hard attention to restrict the area used for classification. \cite{poz20} incorporated attention and reinforcement learning.

Our approach differs in several ways. First, we do not learn an attention mechanism by (i) using a sigmoid function to obtain attention scores and (ii) multiplying these scores with activations from a forward pass, or (iii) using gradient information to adjust weights (of attention and other layers). Instead, we concatenate explanations to a layer. Second, our explanations are "feature and location-based" rather than just "location-based". While some methods use multiple, more or less independent attention maps, our approach involves a single, detailed map derived directly from a single explanation. Third, and most importantly, our training heavily relies on using explanations from correct and incorrect classes. Rather than emphasizing the importance of ``paying attention," we focus more on the goal of reflection, i.e., ``Let's investigate multiple decisions and see what decisions should be made.''

\medskip
\noindent\emph{Gradient Noise:} Adding noise to gradients \cite{nee15,yan20} during training can also be seen as a form of augmentation or regularization. However, adding explanations is generally unrelated to adding noise to gradients. First, we do not change gradients but add input information (in the form of explanations), and second, explanations are not typically simply gradients. However, adding noise to the explanation, i.e. if we augment explanations, might be seen coarsely as altering a function of the gradients if explanations are computed using gradients (e.g., as in GradCAM).

\medskip\noindent
\noindent\emph{Knowledge Transfer:} In \cite{li2019}, explanations (i.e., attention maps) were used to transfer knowledge from a single-label dataset to a multi-label dataset using regularization. While our approach is motivated by self-reflection, it could also be used for combining knowledge from multiple models, known as knowledge distillation \cite{hin15}. However, our approach does not follow the typical student-teacher paradigm \cite{wang20,tan18}, as there is no ``teacher" guiding a student. Instead, self-reflection/self-explanation is used as input to make a decision. Knowledge from one or more networks can also be transferred to another network using input weighing \cite{dhu19}.

\medskip
\noindent\emph{Self-supervised learning:} Some approaches aim to learn without labeled data, such as aligning multiple modalities \cite{de94} or using structural insights of the data \cite{jin20}. These approaches may use features learned from a prediction task to perform a classification task. In contrast, our self-reflective learning approach does not use the explanation as a ``supervision", or label. Instead, the explanation is simply an input to the network.

\medskip
\noindent\emph{Explainability:} The field of explainability is rapidly evolving, but still faces challenges \cite{mes21}. One concern that has been raised is that the intentional use of explanations of "wrong" classes could be seen as a deception attempt \cite{sch20}. Our approach can be used with any explainability method, but is most suitable for methods that explicitly provide explanations for any class for any input, such as GradCAM, LIME, and LRP. Methods like ClaDec \cite{sch22exp} are designed to only explain the actual prediction, so using a randomly generated explanation with such a method is likely to produce inferior results. LRP \cite{bach2015pixel} is a popular method, but has been criticized for being sensitive to mean shift of inputs \cite{kin19} and being dominated by the input \cite{adebayo2018sanity}. LIME \cite{rib16} is computationally demanding, as it requires training a proxy model for each local explanation.

\medskip
\noindent\emph{Self-reflection:} Self-reflection is a process of introspection and critical examination of one's own thoughts and actions. It has been studied in psychology and neuroscience as a means of gaining self-insight and improving decision-making. In technical systems, the concept of self-reflection is largely absent. However, some research has explored ways to incorporate self-reflection into machine learning systems. For example, \cite{tom14} proposed high-level ideas for incorporating self-reflection into systems, while \cite{alt16} used the term to describe mechanisms for altering the relevance of experiences in reinforcement learning. The process of self-reflection has been studied in neuroscience, for example by using MRI scans to identify brain regions involved in self-reflection \cite{joh02}. In psychology, self-reflection has been widely discussed \cite{hix93}, and it has been found that even human self-reflection does not always result in self-insight. This aligns with our findings that "trivial forms" of self-reflection in a system may not lead to improvements.

\section{Discussion and Future Work}
\noindent
\emph{What explanation to use?} 
There is a wide range of explainability methods available, including attribution-based techniques like GradCAM\cite{selvaraju2017grad} and concept-based methods\cite{sch22exp}. GradCAM was one of the methods said to have passed elementary sanity checks that many other methods did not~\cite{adebayo2018sanity}.  GradCAM has also been found to be effective in user studies, by creating explanations that deceive people to perceive the ``wrong'' prediction as true\cite{sch20}. However, it is not clear whether it provides a comprehensive explanation in the sense of showing how relevant input parts interact with each other. This raises questions about the suitability of techniques that aim to explain ML models to humans for self-reflection. This could be a direction for future research in the field of ``explainability for self-reflection" in addition to the existing field of "explainability for humans". Our work focuses on pure self-reflection without any human intervention.

\noindent
\emph{The role of using multiple explanations per input.}
One key aspect of our work was to reflect on explanations for different predictions, not just the most likely one. For gradient-based techniques like GradCAM, this is intuitive:gGradients for the most likely class are already included during training with regular stochastic training. Therefore, adding explanations using the same gradients does not provide new information. Using multiple explanations also serves as a regularizer for the reflective network, preventing the network from relying too much on the explanations and making decisions primarily on them. 

\noindent
\emph{How to incorporate explanations?}
In addition to incorporating explanations through layer concatenation, we suggest that other mechanisms such as attention may also be a promising direction for future research. Also other themes, arising for non-reflective networks such as initialization should be investigated, e.g., whether correlated initialization is helpful or not \cite{sch22cor}.

\noindent
\emph{Self-reflection in humans vs. our work.} While our work is partially inspired by the human capability of self-reflection, our concept of self-reflection is more limited compared to the more comprehensive notion in humans \cite{hix93}. Our concept refers to reflecting on individual decisions in sequence, while humans reflect on their capabilities, actions, and emotions as a whole. Therefore, we do not claim to replicate human capabilities in this regard, similar to other human traits like creativity \cite{bas22}. However, it is possible that a model could also learn from explanations generated by humans. In such a setup, a human would need to explain in terms of the input, i.e., highlight areas of the input that are relevant for decision making, or understand the model's representation and align it with their own explanations for upper layers in the network. 

\noindent
\emph{Data augmentation and computational costs:} Our approach can be seen as a form of data augmentation. Any data augmentation comes with computational costs. In contrast to simple augmentations like image rotations, flipping, etc., computing explanations based on our method require computing gradients of a network and is thus computationally expensive. This is a limitation.

Furthermore, other data (e.g., text) and models (e.g., large scale transformers \cite{sch22f}) could be investigated as well.

\section{Conclusions}

Our study has shown that incorporating data generated by explanation techniques, which promote self-reflection, can significantly improve classifier performance and training efficiency in multiple image classification datasets and convolutional neural network architectures. This is an important result as it demonstrates the potential for machines to learn from their own explanations and engage in a form of reflective thinking, similar to humans. Our approach also opens up new possibilities for data augmentation, by capitalizing on explanations for both the correct and incorrect classes. 

\section*{Declarations}
\begin{itemize}
\item Conflict of interest/Competing interests:
The authors have no competing interests to declare that are relevant to the content of this article.
\item Funding:
The authors did not receive support from any organization for the submitted work.
\item Authors' contributions: Johannes Schneider contributed through the idea, conceptualization, implementation and writing up to the first draft. Michalis Vlachos supported related work, introduction and improvement of the manuscript including all revisions.
\end{itemize}

\bibliographystyle{spmpsci}
\bibliography{refs}
\end{document}